\documentclass[10pt,twocolumn,letterpaper]{article}

\usepackage{cvpr}
\usepackage{times}
\usepackage{epsfig}
\usepackage{graphicx}
\usepackage{amsmath}
\usepackage{adjustbox}
\usepackage{amssymb}
\usepackage{booktabs}
\newtheorem{myProblem}{Problem}
\graphicspath{{figures/}}

\usepackage[breaklinks=true,bookmarks=false]{hyperref}

\cvprfinalcopy 

\def\cvprPaperID{****} 

\begin{document}

\title{Sparse Subspace Clustering via Diffusion Process}

\author{Qilin Li, Ling Li, Wanquan Liu\\
Curtin University, Perth, Australia\\
{\tt\small kylinlovesummer@gmail.com}
}

\maketitle

\begin{abstract}
Subspace clustering refers to the problem of clustering high-dimensional data that lie in a union of low-dimensional subspaces. State-of-the-art subspace clustering methods are based on the idea of expressing each data point as a linear combination of other data points while regularizing the matrix of coefficients with $\ell_1$, $\ell_2$ or nuclear norms for a sparse solution. $\ell_1$ regularization is guaranteed to give a subspace-preserving affinity (i.e., there are no connections between points from different subspaces) under broad theoretical conditions, but the clusters may not be fully connected. $\ell_2$ and nuclear norm regularization often improve connectivity, but give a subspace-preserving affinity only for independent subspaces. Mixed $\ell_1$, $\ell_2$ and nuclear norm regularization could offer a balance between the subspace-preserving and connectedness properties, but this comes at the cost of increased computational complexity. This paper focuses on using $\ell_1$ norm and alleviating the corresponding connectivity problem by a simple yet efficient diffusion process on subspace affinity graphs. Without adding any tuning parameter , our method can achieve state-of-the-art clustering performance on Hopkins 155 and Extended Yale B data sets.
\end{abstract}

\section{Introduction}
Many computer vision problems, such as image compression \cite{hong2006multiscale}, motion segmentation \cite{rao2010motion} and face clustering \cite{ho2003clustering}, deal with high-dimensional data. The high-dimensionality of the data not only increases the computational time and memory requirements of algorithms, but also decreases their performance due to the noise effect and insufficient number of samples with respect to the ambient space dimension, commonly referred to as the ``curse of dimensionality" \cite{Bellman1957Dynamic}. However, even though data are high-dimensional, their intrinsic dimension is often much smaller than the dimension of the ambient space, which has motivated the development of a number of techniques for finding a low-dimensional representation of a high-dimensional data set. Conventional techniques, such as Principal Component Analysis (PCA), assume that the data is drawn from a single low-dimensional subspace of the high-dimensional space. In practice, however, such high-dimensional data usually lie close to multiple low-dimensional subspaces corresponding to several classes or categories to which the data belong. In these scenarios, the task of clustering a high-dimensional data set into multiple classes becomes to the task of assigning each data point to its own subspace and recovering the underlying low-dimensional structure of the data, a problem known in the literature as \emph{subspace clustering} \cite{vidal2010tutorial}.

In machine learning and computer vision communities, existing subspace clustering methods can be divided into four main categories, including algebraic methods \cite{costeira1998multibody,vidal2005generalized}, iterative methods \cite{tseng2000nearest,lu2006combined}, statistical methods \cite{ma2007segmentation,rao2008motion}, and spectral clustering-based methods \cite{zhang2012hybrid,goh2007segmenting,chen2009spectral,elhamifar2013sparse,liu2010robust}. Among them, spectral-clustering based methods have become extremely popular due to their simplicity, theoretical correctness, and empirical success. These methods generally divide the problem into two steps: 1) Constructing an affinity matrix based on certain model and 2) applying spectral clustering to the affinity matrix. In this paper, we focus on the first step since the success of spectral clustering highly depends on having an appropriate affinity matrix.

State-of-the-art methods for constructing the affinity matrix in terms of subspace clustering are based on the \emph{self-expressiveness property} of the data \cite{elhamifar2009sparse}, i.e., each data point in a union of subspaces can be efficiently reconstructed by a linear combination of all other data points: $x_j=\sum_{i\neq j}c_{ij}x_i$, where the coefficient $c_{ij}$ is used to define the affinity between points $i$ and $j$ as $w_{ij}=|c_{ij}| + |c_{ji}|$. However, this leads to an ill-posed problem with many possible solutions. To deal with this issue, the principle of \emph{sparsity} is invoked. Specifically, every point is expressed as a sparse linear combination of all other data points by minimizing certain norm of coefficient matrix. This problem can then be written as:

\begin{equation}
\label{optimizationl1}
\min_C||C||, \quad  s.t. \quad  X=XC, \quad  diag(C)=0,
\end{equation}
where $X=[x_1,...,x_N]$ is the data matrix, $C=[C_1,...,C_N]$ is the coefficient matrix, $||\cdot||$ is a properly chosen regularizer.

The main difference among state-of-the-art methods lies in the choice of the regularizer. In Sparse Subspace Clustering (SSC) \cite{elhamifar2009sparse}, the $\ell_1$ norm is used for $||\cdot||$ as a convex relaxation of the $\ell_0$ norm to promote the sparseness of $C$. While under broad theoretical conditions \cite{elhamifar2013sparse,you2015geometric} the sparse representation produced by SSC is guaranteed to be subspace-preserving (i.e.,$c_{ij}\neq 0$ only if $x_i$ and $x_j$ are in the same subspace), the affinity graph however may lack connectedness \cite{} (i.e., data points from the same subspace may not form a connected component of the affinity graph due to the sparseness of the connections). In Low-Rank Representation (LRR) \cite{liu2010robust} and Low-Rank Subspace Clustering \cite{favaro2011closed}, the nuclear norm $||\cdot||_\ast$ is adopted as a convex relaxation of the rank function. One benefit of nuclear norm is that the coefficient matrix is generally dense, which alleviates the connectivity issue of sparse representation based methods. However, the representation matrix is known to be subspace preserving only when the subspaces are independent, which significantly limits its applicability.

\begin{figure*}[t]
\begin{center}
   \includegraphics[width=0.8\linewidth]{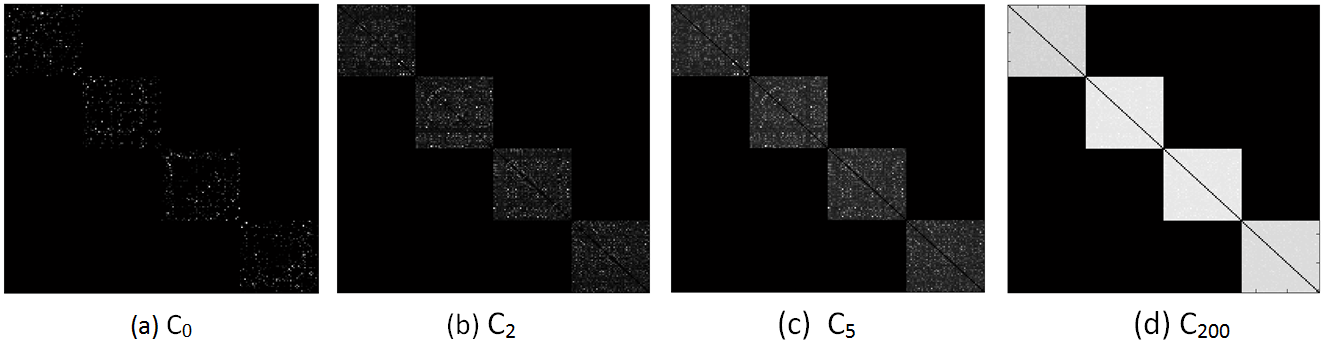}
\end{center}
   \caption{Visualization of affinity matrices obtained by different methods: (a) $C_0$ is the original coefficient matrix obtained by SSC($\ell_1$ norm), (b)$\backsim$(d) $C_t$ are the matrices after $t$ steps of diffusion process applied on $C_0$. Clearly, as the diffusion process goes on, the block-diagonal structure of the affinity matrix becomes evident.}
\label{affinityExample}
\end{figure*}

In this paper, we propose to use $\ell_1$ norm for sparsity constraint to best retain the subspace preserving property, and adopt a diffusion process on subspace of affinity graph to mitigate the connectedness problem. Specifically, the fist step is to learn a sparse affinity matrix by applying $\ell_1$ norm on optimization problem of (\ref{optimizationl1}). Diffusion process is then adopted to spread the affinity values through the entire graph built upon the affinity matrix. Such a process is interpretable as a random walk on the graph, where a so-called transition (affinity) matrix defines probabilities for walking from one node to a neighboring node. One remarkable advantage here is, since the affinity matrix learned by $\ell_1$ norm is subspace preserving, the random walk on this matrix is guaranteed to be subspace constrained. Therefore, the connectivity within subspace is significantly enhanced while the subspace preserving property remains unaltered. An illustrative example is given in Figure \ref{affinityExample}. It clearly demonstrates that based on the sparse affinity matrix obtained by $\ell_1$ norm, the proposed method could evidently improve the affinity within subspaces while the sparsity between subspaces is retained, yielding affinity matrix with exactly block-diagonal structure.

\section{Related Work}
\subsection{Subspace clustering}
State-of-the-art subspace clustering methods are based on the self-expressiveness model. The main difference among those methods lies in the choice of the regularizer on the coefficient matrix. While different regularizers possess their own advantages and drawbacks, \cite{wang2013provable,panagakis2014elastic,you2016oracle} propose to use mixed norms. For example, the low-rank sparse subspace clustering (LRSSC) method \cite{wang2013provable} combines $\ell_1$ and nuclear norm regularizer:

\begin{equation}
\label{mixl1nuclear}
||C||_\ast+\lambda||C||_1,
\end{equation}
where $\lambda$ controls the trade-off between the two regularizers. Likewise, \cite{panagakis2014elastic,you2016oracle} propose to use a mixed $\ell_1$ and $\ell_2$ norm given by

\begin{equation}
\label{mixl1l2}
\lambda||C||_1+\frac{1-\lambda}{2}||C||_2,
\end{equation}
where $\lambda$ plays a trade-off role between the two norms. These methods basically attempt to bridge the gap between the subspace preserving and connectedness properties by the trade-off of different norms.

\cite{feng2014robust} proposes to explicitly impose a block-diagonal constraint by fixing the rank of Laplacian matrix. One benefit of this approach is that it can be applied to all the affinity construction methods straightforwardly. The Structured Sparse Subspace Clustering (SSSC) \cite{li2015structured} integrates the two stages, affinity learning and spectral clustering, into one unified optimization framework. Their observation is that the clustering results can help the self-expressiveness model to yield a better affinity matrix.

\subsection{Diffusion processes}
Diffusion process is widely used in the field of retrieval \cite{yang2009locally,bai2010learning,egozi2010improving,wang2012affinity,donoser2013diffusion}, in which the task is retrieving the most similar instances to a provided query element from a potentially large database. Conventional approaches are usually based on analyzing pairwise affinity/distance values which are directly used to rank the most similar elements afterwards. Such approaches has the main limitation that the structure of the underlying data manifold is completely ignored. For this reason, instead of considering pairwise affinity individually, diffusion process is adopted to derive context sensitive measures, and it has shown to be an indispensable tool for improving retrieval performance \cite{donoser2013diffusion}.

Diffusion process is generally start with a affinity matrix $A_{N\times N}$. The first step is to interpret the matrix $A$ as a undirected graph $G=(V,E)$, consisting of $N$ nodes $\emph{v}_i\in V$, and edges $\emph{e}_{ij}\in E$ that link nodes to each other. The edge weights are fixed to the affinity values $A_{ij}$. Diffusion process then spreads the affinity values through the entire graph based on the defined edge weights.

Diffusion process can be interpreted as a Markov random walk on a graph $G=(V,E)$. To this end, we first define the transition matrix of random walk as

\begin{equation}
\label{transitionMatrix}
P=D^{-1}A,
\end{equation}
where $D$ is a diagonal matrix with $D_{ii}=\sum_{k=1}^nA(i,k)$. Obviously, $P$ is a row-stochastic matrix, containing the transition probabilities for a random walk in the corresponding graph. With a simple undate rule $A_{t+1}=A_tP$, the affinity matrix after $t$ steps of random walks can be obtained by

\begin{equation}
\label{randomWalk}
A_t=AP^t.
\end{equation}

The random walk model was later extended to one of the most successful retrieval methods, the Google PageRank system \cite{page1999pagerank}. The standard random walk is modified, and at each time step $t$ a random walk step is done with probability $\alpha$, whereas a random jump to an arbitrary node is made with probability $(1-\alpha)$. This leads to following update strategy

\begin{equation}
\label{randomWalk}
A_{t+1}=\alpha A_tP+(1-\alpha)Y,
\end{equation}
where $Y$ is probabilities of randomly jumping, which enables personalization for individual web user. Similar method was proposed in \cite{zhou2004ranking}, namely Ranking on Manifolds. The different is the slightly adapted transition matrix $P=D^{-1/2}AD^{-1/2}$.

The drawback of the diffusion process mentioned above is that the process is applied to the entire graph, which can be heavily influenced by noisy edges. Current state-of-the-art methods \cite{yang2011affinity} restricted the diffusion process to the $K$ nearest neighbor (KNN) graph. Given affinity matrix $A$, this method sets $A_{ij}=A_{ij}$ only if $x_j \in KNN(x_i)$, otherwise $A_{ij}=0$, and comes up with a new update strategy as

\begin{equation}
\label{lcdp}
A_{t+1}=PA_tP^T.
\end{equation}
While this approach consistently yields state-of-the-art performance in the field of retrieval, it is observed that in practice one needs to set neighborhood $K$ manually and the performance is sensitive to the choice of $K$ \cite{donoser2013diffusion}.

\textbf{Paper Contributions.} In this paper, we exploit diffusion process to mitigate connectedness problem of $\ell_1$ norm in terms of subspace clustering. To the best of our knowledge, this is the first attempt to adopt the idea of diffusion to this field. Since the idea of using $\ell_1$ norm for subspace clustering is originally from Sparse Subspace Clustering (SSC), we refer our method as Diffusion-based Sparse Subspace Clustering (DSSC). Our main contributions can be summarized as:
\begin{enumerate}
  \item For subspace clustering, instead of adding different norms to balance the subspace preserving and connectivity properties, we come up with using $\ell_1$ norm and alleviating the corresponding connectivity problem by a simple yet efficient diffusion process. Without adding any tuning parameter, the widely existing gap between the two properties is well bridged.
  \item From the diffusion point of view, we show that instead of choosing $KNN$ neighbors based on Euclidean distance, the sparse property of $\ell_1$ norm provides manifold-aware neighborhood construction, which are locally constrained in corresponding subspaces. Moreover, the tuning parameter $k$ of original diffusion is eliminated.
  \item We present experiments on both synthetic data sets and real computer vision data sets that demonstrate the superiority of the proposed method compared to other state-of-the-art methods.
\end{enumerate}

\section{Diffusion based Sparse Subspace Clustering}
Before introducing the proposed approach, we first formulate the addressed problem in this paper.

\begin{myProblem}
Given a collection of data points $\{x_j \in \mathbb{R}^D \}_{j=1}^N$  drawn from an unknown union of k subspaces $\{S_i\}_{i=1}^k$ of unknown dimensions $d_i=dim(S_i),0<d_i<D,i=1,...,k$. The goal is to segment these points into their corresponding subspaces.
\end{myProblem}

Sparse Subspace Clustering (SSC) attempts to solve the problem based on the so-called self-expressiveness model, which states that each data point can be expressed as a linear combination of all other data points, i.e., $X=XC+E$, where $C$ is the coefficient matrix and $E$ is the matrix of error (noises or outliers). In principle, this leads to an ill-posed problem with many possible solutions. Thus, the sparsity constraint is invoked by $\ell_1$ minimization, leading to the following optimization problem

\begin{equation}
\label{optimizationl0}
\min_C||C||_1+||E||_\ell \quad  s.t. \quad  X=XC+E, \quad  diag(C)=0,
\end{equation}
where the Frobenius norm or $\ell_1$ norm is used for $||\cdot||_\ell$ to handle noise or outliers. The SSC algorithm proceeds by solving the optimization problem in (\ref{optimizationl0}) using the ADMM method. The optimal coefficient $C$ is then used to define an affinity matrix $|C|+|C^T|$. The segmentation of the data is finally obtained by applying spectral clustering to the normalized Laplacian.

While SSC works well in practice, one possible drawback is $\ell_1$ finds the sparse representation of each data point individually. In the case of clean data, SSC is guaranteed to be subspace-preserving, i.e., there are no connections between points from different subspaces. However, the within-subspace connections are usually sparse, i.e., $C_{ij}$ could be zero even $x_i$ and $x_j$ are in the same subspace. It is not a problem as long as the connections are still subspace-preserving. But in the case of noisy data, there is no theoretical guarantee that the nonzero coefficients correspond to points in the same subspace. Imaging there are connections between points from different subspaces, spectral clustering cannot be able to appropriately cut the graph, as shown in Figure \ref{graphCut2}.

Diffusion process is capable to deal with this issue by exploiting the contextual affinities. Given the affinity matrix $W=|C|+|C^T|$, where $C$ is obtained by solving (\ref{optimizationl0}), diffusion process is encoded into computing the power of the affinity matrix, which is

\begin{equation}
W_t=W^t,
\end{equation}
where $t$ corresponds $t$ steps of diffusion process. Obviously, such a process is sensitive to the step $t$. In order to make the diffusion process independent from $t$, we consider the accumulation of all $t$. Thus, the diffusion process is

\begin{equation}
\label{classicDiffusion}
W_t=\sum_{i=0}^tW^i.
\end{equation}

We assume that $W$ is nonnegative and the sum of each row is smaller than one. A matrix $W$ that satisfies these requirements can be easily constructed from a stochastic matrix. Note that the absolute values of the eigenvalues is bounded by the maximum of the rowwise sums. Therefore, the maximum of the absolute values of the eigenvalues of $W$ is smaller than one. Consequently, (\ref{classicDiffusion}) converges to a fixed and nontrivial solution given by $\lim_{t\rightarrow \infty}W_t=(I-W)^{-1}$, where $I$ is the identity matrix.

To further incorporate the contextual affinity, it is shown in \cite{yang2011affinity} that the diffusion process on higher order tensor product graph is promising for revealing the intrinsic relation between data points. Given graph $G=(V,E)$ constructed from affinity matrix $W$, the tensor product graph is defined as the Kronecker product of original graph, $\mathbb{G}=G\bigotimes G$. The corresponding affinity matrix is $\mathbb{W}=W\bigotimes W$. In particular, we have

\begin{equation}
\mathbb{W}(\alpha,\beta,i,j)=W(\alpha,\beta)\cdot W(i,j)=w_{\alpha,\beta}\cdot w_{i.j}.
\end{equation}
Thus, if $W\in \mathcal{R}^{n\times n}$, then $\mathbb{W}=W\bigotimes W\in \mathcal{R}^{nn\times nn}$.

The diffusion process is then defined on the higher order tensor as

\begin{equation}
\label{tpgDiffusion}
\mathbb{W}_t=\sum_{i=0}^t\mathbb{W}^i.
\end{equation}

Since the sum of each row of $W$ is smaller than 1, we have

\begin{equation}
\sum_{\beta j}\mathbb{W}(\alpha,\beta,i,j)=\sum_{\beta j}w_{\alpha \beta}w_{ij}=\sum_{\beta}w_{\alpha \beta}\sum_jw_{ij}<1.
\end{equation}

As is the case for (\ref{classicDiffusion}), the process (\ref{tpgDiffusion}) also converges to a fixed and nontrivial solution,

\begin{equation}
\lim_{t\rightarrow \infty}\mathbb{W}_t=\lim_{t\rightarrow \infty}\sum_{i=0}^t\mathbb{W}^i=(I-\mathbb{W})^{-1}
\end{equation}

Since our goal is to learn a new affinity matrix $W^*$ of size $n\times n$, it is defined as

\begin{equation}
W^*=vec^{-1}((I-\mathbb{W})^{-1}vec(I)),
\end{equation}
where $I$ is the identity matrix and $vec$ is an operator that stacks the columns of a matrix one after the next into a column vector. The inverse of $vec$ is denoted as $vec^{-1}$.

While the tensor graph provides adequate underlying structure of the data, it is impractical for large scale problems due to the demand of high storage and computing cost. Therefore, we use an iterative algorithm for the diffusion process on tensor graph. First, we define $A_1=W$ and the update strategy as

\begin{equation}
\label{tpgIteration}
A_{t+1}=WA_tW^T+I,
\end{equation}
where $I$ is the identity matrix. The diffusion process becomes the iteration of (\ref{tpgIteration}) until convergence. To prove the convergence of (\ref{tpgIteration}), we first transform (\ref{tpgIteration}) to

\begin{align}
\nonumber
A_{t+1}&=WA_tW^T+I=W(WA_{t-1}W^T+I)W^T+I \\ \nonumber
       &=W^2A_{t-1}(W^T)^2+WIW^T+I= ... \\ \nonumber
       &=W^tW(W^T)^t+W^{t-1}I(W^T)^{t-1}+...+I \\
       &=W^tW(W^T)^t+\sum_{i=0}^{t-1}W^iI(W^T)^i.
\end{align}

Since we assume the sum of each row of $W<1$, we have $\lim_{t\rightarrow \infty}W^tW(W^T)t=0$, consequently,

\begin{equation}
\lim_{t\rightarrow \infty}A_{t+1}=\lim_{t\rightarrow \infty}\sum_{i=0}^{t-1}W^iI(W^T)^i.
\end{equation}

As shown in \cite{yang2011affinity}, it can be proven by induction that

\begin{equation}
\lim_{t\rightarrow \infty}\sum_{i=0}^{t-1}W^iI(W^T)^i=vec^{-1}((I-\mathbb{W})^{-1}vec(I)).
\end{equation}
So we have

\begin{equation}
\lim_{t\rightarrow \infty}A_{t+1}=vec^{-1}((I-\mathbb{W})^{-1}vec(I)).
\end{equation}
Hence, the iterative algorithm (\ref{tpgIteration}) yields the same affinities as the diffusion process on tensor graph.

\subsection{A graph view of the diffusion process}
With $\ell_1$ minimization, SSC seeks sparse representation for each data point individually. The corresponding affinity graph reveals the pairwise affinity as the ``shortest path" between them, which is susceptible to noise. The proposed DSSC derives the affinity by considering the ``volume of paths" through a diffusion process. One remarkable invention is, with $\ell_1$ norm, the ``volume of paths" are restricted in subspaces, resulting in the enhancement of within-subspace affinity and constant of between-subspace affinity.

\begin{figure}[t]
\begin{center}
   \includegraphics[width=0.7\linewidth]{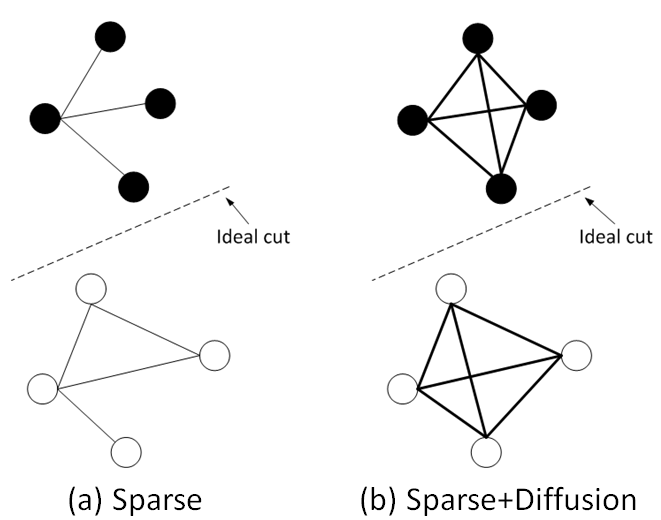}
\end{center}
   \caption{An example of affinity graph for clean data, obtained by different methods: (a) sparse, (b) sparse with diffusion.}
   \label{graphCut1}
\end{figure}

Figure \ref{graphCut1} shows that in the case of clean data, SSC is guaranteed to be subspace-preserving, i.e., there are no connections between subspaces. Thus, spectral clustering can properly cut the graph. However, due to the sparse property of $\ell_1$ norm, it is very likely that two points in the same subspace are not connected. While in such a case, the diffusion process can accurately complete the graph for each subspace, leading to more robust subspace graphs.

\begin{figure}[t]
\begin{center}
   \includegraphics[width=0.7\linewidth]{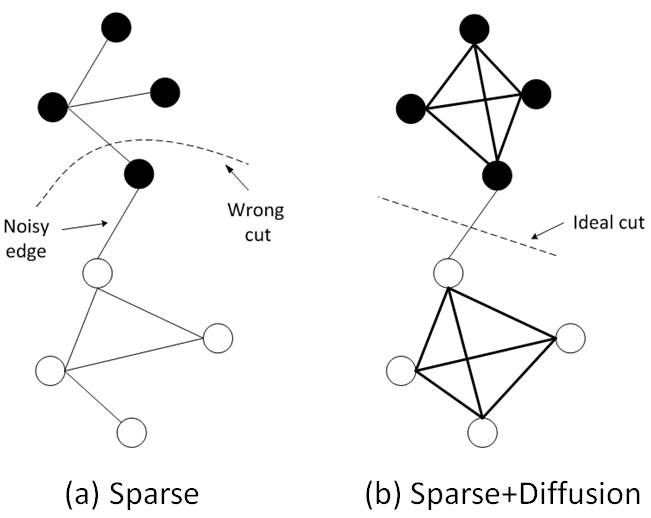}
\end{center}
   \caption{An example of affinity graph for noisy data, obtained by different methods: (a) sparse, (b) sparse with diffusion.}
   \label{graphCut2}
\end{figure}

In the case of noisy data, the advantage of diffusion becomes more significant. As shown in Figure \ref{graphCut2}, when there are noisy edges between subspaces, spectral clustering may not be able to find the ideal cut of the graph. As the pairwise affinity is individually computed in SSC, it may be difficult to distinguish ``noisy edge" with ``real edge". While in DSSC, due to the ability of combining contextual information, the within-subspace affinities are evidently enhanced so that the noisy edges can be properly cut by spectral clustering. It should be noticed that real world data are usually noisy, which explains the significant improvements on real world data sets, shown in Section \ref{Hopkins155} and Section \ref{YaleB}.

\subsection{A random walk view of the diffusion process}
As we mentioned, the proposed diffusion process can be interpreted as Markov random walks \cite{jaakkola2002partially}, where the transition (affinity) matrix defines probabilities for walking from one node to a neighboring node. The crucial observation is that considering a affinity graph, there will be many connections within a subspace, and fewer connections between subspaces. Therefore, if we start a walker at one node and then randomly travel to a connected node, the random walker is more likely to stay within the subspace than travel between different subspaces. Intuitively, diffusion process can be imaged as random walks starting at every node on the graph.  As the random walks go on,  the probabilities of traveling between nodes within subspace will increase, while the probabilities between subspaces decrease.

Given affinity matrix $A$, one can easily build a stochastic transition matrix $P$ by (\ref{transitionMatrix}). The probability distribution of the positions of a random walk, starting at node $j$, is given by the $j$-th row of the transition matrix. After $t$ time steps, the probability of a random walk starting at node $j$ at time 0, to be at node $i$ at time $t$ is thus,

\begin{equation}
p_t(i|j)=P^t_{i,j}.
\end{equation}
Under mild conditions \cite{chung1997spectral}, we obtain in this way an ergodic Markov process with a single stationary distribution $\pi=(\pi_1,...,\pi_n)$, where $\pi_i=d_i/vol(V)$, $d_i$ is the degree of node $i$, $V$ is the set of all nodes. It is easy to verify that this distribution is a right-eigenvector of the $t$-step transition matrix (for every t), e.g. $P^T\pi=\pi$, since $\pi_i=\sum P_{i,j}\pi_j$.

As we know, there is a tight relation between random walks and spectral segmentation \cite{meila2001random}. We take the Normalized Cut \cite{shi2000normalized} (NCut) as an example of spectral segmentation. The goal of NCut algorithm is to segment an image into two disjoint parts by minimizing

\begin{equation}
\label{NCut}
NCut(A,\overline{A})=\Bigg(\frac{1}{vol(A)}+\frac{1}{vol(\overline{A})}\Bigg)\sum_{i\in A, j\in \overline{A}}A_{ij}.
\end{equation}
As derived by \cite{meila2001random}, the objective function of NCut (\ref{NCut}) can be expressed in the framework of random walk as follows. For two disjoint subsets $A$, $B \subset V$, assume we run a random walk starting with $X_0$ in the stationary distribution $\pi$. We define

\begin{equation}
P(B|A)=P(X_1\in B|X_0 \in A)
\end{equation}
as the probability of the random walk transition from set $A$ to set $B$. First of all it can be observed that

\begin{align}
\nonumber
P(X_0\in A, X_1\in B)&=\sum_{i\in A, j\in B}P(X_i,X_j) \\ \nonumber
                     &=\sum_{i\in A, j\in B}\pi_ip_{ij} \\ \nonumber
                     &=\sum_{i\in A, j\in B}\frac{d_i}{vol(V)}\frac{a_{ij}}{d_i} \\
                     &=\frac{1}{vol(V)}\sum_{i\in A, j\in B}a_{ij}.
\end{align}
With the Bayes' Rules, we have the posterior probability

\begin{align}
\nonumber
P(B|A)&=\frac{P(X_0\in A, X_1\in B)}{P(X_0\in A} \\ \nonumber
       &=\Bigg(\frac{1}{vol(V)}\sum_{i\in A, j\in B}a_{ij}\Bigg)\Bigg(\frac{vol(A)}{vol(V)}\Bigg)^{-1} \\
       &=\frac{1}{vol(A)}\sum_{i\in A, j\in B}a_{ij}.
\end{align}
Follow this, the definition of NCut can be written as

\begin{equation}
\label{NCut}
NCut(A,\overline{A})=P(\overline{A}|A)+P(A|\overline{A}).
\end{equation}
While originally the objective of NCut algorithm is to find those regions that the between connections are minimized, it can be interpreted as to find regions in which the probabilities of random walkers escape from these regions are low, with the theory of random walk.

Denoting the transition matrix after $t$ time steps random walk as $P_t$, we know that $P_t(\overline{A}|A)<P(\overline{A}|A)$ while $P_t(A|A)>P(A|A)$. Let $NCut^*$ be the NCut criterion on the transition matrix after random walk, the following holds,

\begin{equation}
\label{NCut}
NCut^*(A,\overline{A})=P_t(\overline{A}|A)+P_t(A|\overline{A})<NCut(A,\overline{A}),
\end{equation}
which explains why random walk (diffusion process) can improve the performance of spectral segmentation.

\section{Experiments}
Experiments are demonstrated in this section. We evaluate the proposed DSSC approach on a synthetic data set, a motion segmentation data set, and a face clustering data set to validate its effectiveness.

\textbf{Experimental Setup.} Since the proposed DSSC is built upon the standard SSC \cite{elhamifar2013sparse}, we keep all settings in DSSC the same as in SSC. As for diffusion process, since it is guaranteed to converge, we set the iteration to 200 for all experiments. The performance is validated by clustering error, which is measured by

\begin{equation}
clustering\ \ error = \frac{\#\ of\ missclassified\ \ points}{total\ \#\ of\ points}.
\end{equation}
\subsection{Experiments on synthetic data}
\textbf{Data Generation}. We construct 5 subspaces $\{S\}_{i=1}^5\subset \mathbb{R}^{100}$ whose bases $\{U_i\}_{i=1}^5$ are obtained by $U_i=T_iU, 1\leq i \leq 5$, where $U$ is a random matrix of dimension 100$\times$ 5 and $T_i^{100\times 100}$ is a random rotation. We sample 50 data points from each subspace by $X_i=U_iQ_i$, where the entries of $Q_i \in \mathbb{R}^{5,50}$ are i.i.d. samples from a standard Gaussian. Some data vectors x are then randomly chosen to corrupt by Gaussian noise with zero mean and variance 0.3$||x||$.

\begin{table*}[t]
  \caption{Clustering Errors on Synthetic Dataset.}
  \label{resultsSynthetic}
  \centering
  \begin{adjustbox}{width=0.8\textwidth}
  \begin{tabular}{cccccccccccc}
  \toprule
  Corruptions & 0 & 10\% & 20\% & 30\% & 40\% & 50\% & 60\% & 70\% & 80\% & 90\% & 100\% \\
  \hline
  SSC   & 0 & 2.88 & 6.90 & 13.26 & 18.66 & 24.62 & 30.16 & 34.42 & 38.80 & 40.20 & 43.00 \\
  DSSC  & \textbf{0} & \textbf{1.60} & \textbf{4.80} & \textbf{9.74} & \textbf{14.40} & \textbf{17.64} & \textbf{22.80} & \textbf{28.98} & \textbf{32.80} & \textbf{36.00} & \textbf{39.74} \\
  \bottomrule
  \end{tabular}
  \end{adjustbox}
\end{table*}

Experimental results are presented in Table \ref{resultsSynthetic}. It can be observed that the proposed DSSC consistently outperforms SSC. In the case of clean data (no corruptions), both SSC and DSSC achieve perfect clustering with error rates equal to 0. As the corruptions increase, so do the chance of adding noise edges into the corresponding affinity graph. In this case, spectral clustering cannot find the appropriate cut of the graph in SSC. While in DSSC, those noisy edges are well detected due to the within subspaces edges are significantly enhanced by the diffusion process, as demonstrated in Figure \ref{graphCut1} and Figure \ref{graphCut2}.


\subsection{Experiments on Motion Segmentation}\label{Hopkins155}
\begin{figure}[t]
\begin{center}
   \includegraphics[width=1\linewidth]{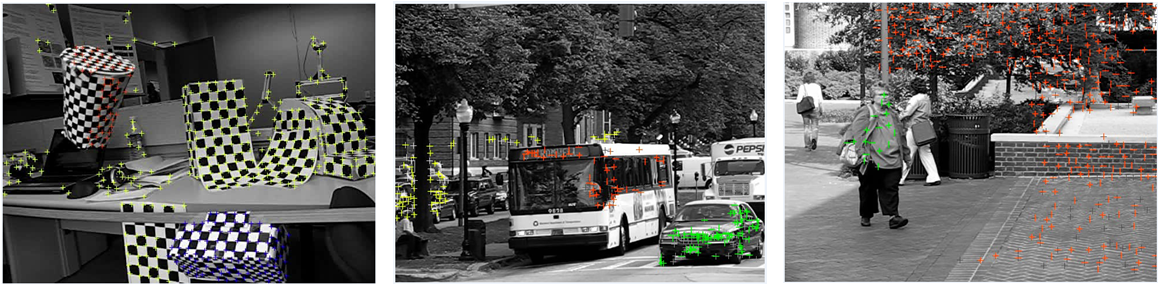}
\end{center}
   \caption{Example frames from videos in the Hopkins 155 \cite{tron2007benchmark}.}
   \label{exampleHopkins155}
\end{figure}

Motion segmentation refers to the problem of segmenting a video sequence of multiple rigidly moving objects into multiple spatiotemporal regions that correspond to different motions in the scene (see Figure \ref{exampleHopkins155}). This problem is often solved by extracting and tracking a set of feature points through all frames of the video. Each data point, which is also called a feature trajectory, corresponds to a vector obtained by stacking all feature points. Under the affine projection model, all feature trajectories associated with a single rigid motion lie in an affine subspace of dimension at most 3 \cite{elhamifar2013sparse}. Therefore, motion segmentation reduces to clustering of these trajectories in a union of subspaces.

We evaluate the proposed DSSC algorithm with other state-of-the-art subspace clustering methods, i.e., SSC \cite{elhamifar2013sparse}, SSSC \cite{li2015structured}, LRR \cite{liu2010robust}, LSR \cite{lu2012robust}, BDSSC \cite{feng2014robust}, LRSC \cite{favaro2011closed}, on the Hopkins 155 motion segmentation data set \cite{tron2007benchmark} for the multi-view affine motion segmentation. It consists of 155 video sequences, where 120 of the videos have two motions and 35 of the videos have three motions. We evaluate average performance on three cases: 2 motions, 3 motions, and all. Experimental results are presented in Table \ref{resultsHopkins}. The result for LRSC is cited from \cite{favaro2011closed}, while the others are cited from \cite{li2015structured}. Note that the proposed DSSC achieves the best performances on all cases.

\begin{table*}[t]
  \caption{Motion Segmentation Errors(\%) on Hopkins 155 Dataset.}\label{resultsHopkins}
  \centering
  \begin{adjustbox}{width=0.8\textwidth}
  \begin{tabular}{cccccccc}
  \toprule
  Methods & LRR \cite{liu2010robust} & LRSC \cite{favaro2011closed} & LSR \cite{lu2012robust} & BDSSC \cite{feng2014robust} & SSC \cite{elhamifar2013sparse}  & SSSC\cite{li2015structured} & DSSC \\
  \hline
  2 motions    & 3.76 & 3.69 & 2.20 & 2.29 & 1.95 & 1.94 & \textbf{1.68} \\
  3 motions  & 9.92 & 7.69 & 7.13 & 4.95 & 4.94 & 4.92 & \textbf{4.64}   \\
  All   & 5.15 & 4.59 & 3.31 & 2.89 & 2.63 & 2.61 & \textbf{2.35} \\
  \bottomrule
  \end{tabular}
  \end{adjustbox}
\end{table*}

\subsection{Experiments on Face Clustering}\label{YaleB}
Given face images of multiple subjects acquired with a fixed pose and varying illumination, we consider the problem of clustering images according to their subjects. It has been shown that, under the Lambertian assumption, images of a subject with a fixed pose and varying illumination lie close to a linear subspace of dimension 9 \cite{basri2003lambertian}. Thus, face clustering can be also considered as a subspace clustering problem, where each subject lies in a 9D subspace.

We evaluate the clustering performance of the proposed DSSC as well as other state-of-the-art methods on the Extended Yale B data set \cite{georghiades2001few}. It contains 2,414 frontal face images of 38 subjects, with 64 images per subject acquired under different illumination conditions. In our experiments, we follow the same settings introduced in \cite{elhamifar2013sparse}. It should be noticed that the Extended Yale B data set is more challenging for subspace segmentation than the Hopkins 155 data set due to the heave noise, high-dimensional space, and large number of subspace in the data.

Experimental results are presented in Table \ref{resultsYaleB}. The results are directly cited from \cite{li2015structured}, which is a fair comparison as we use exactly same experimental settings. It can be observed that the proposed DSSC performs the best results on all cases. Note that on this more challenging data set, DSSC achieves significant improvements compared to the state-of-the-art methods. It is worth mentioning that these significant improvements are achieved by a parameter-free diffusion process, while other methods usually add in tuning parameters for flexibility.

\begin{table*}[t]
  \caption{Clustering Errors (\%) on the Extended Yale B Dataset.}\label{resultsYaleB}
  \centering
  \begin{adjustbox}{width=0.8\textwidth}
  \begin{tabular}{cccccccc}
  \toprule
  Methods & LRR \cite{liu2010robust} & LRSC \cite{favaro2011closed} & LSR \cite{lu2012robust} & BDSSC \cite{feng2014robust} & SSC \cite{elhamifar2013sparse}  & SSSC\cite{li2015structured} & DSSC \\
  \hline
  2 subjects    & 6.74 & 3.15 & 6.72 & 3.90 & 1.87 & 1.27 & \textbf{0.61} \\
  3 subjects  & 9.30 & 4.71 & 9.25 & 17.70 & 3.35 & 2.71 & \textbf{1.25}   \\
  5 subjects  & 13.94 & 13.06 & 13.87 & 27.50 & 4.32 & 3.41 & \textbf{2.80}   \\
  8 subjects  & 25.61 & 21.25 & 25.98 & 33.20 & 5.99 & 4.15 & \textbf{4.04}   \\
  10 subjects  & 29.53 & 29.58 & 28.33 & 39.53 & 7.29 & 5.16 & \textbf{4.84}   \\
  \bottomrule
  \end{tabular}
  \end{adjustbox}
\end{table*}

\section{Conclusion}
In this work, we investigated diffusion process for sparse subspace clustering and proposed a new subspace clustering method, namely DSSC. Specifically, after we obtained affinity matrix by $\ell_1$ minimization, we adopted a diffusion process to spread the affinity value. With the subspace-preserving property of $\ell_1$ norm, such a diffusion process is remarkably constrained within each subspaces, yielding enhanced within-subspace affinity and unaltered between-subspace affinity. Moreover, we explained the diffusion process in the views of graph and random walk, and gave theoretical justifications on how does the diffusion process improve the performance of spectral clustering. Extensive experiments verified that our proposed diffusion based sparse subspace clustering method, without adding in tuning parameter, could significantly improve the state-of-the-art performance.
{\small
\bibliographystyle{ieee}
\bibliography{egbib}
}

\end{document}